\documentclass[oneside,a4paper,11pt,explicit]{book}
\usepackage[utf8]{inputenc}
\usepackage[sedes]{kultem}
\usepackage[english]{babel}
\usepackage{amsmath}

\usepackage[accsupp]{axessibility}  
\usepackage[hidelinks]{hyperref}

\usepackage{lipsum}
\usepackage{listings}
\usepackage{natbib}
\usepackage{blindtext} 
\usepackage{epigraph} 
\usepackage{graphicx}
\usepackage{subfig} 
\usepackage{bm} 
\usepackage{tabularx} 
\usepackage{xcolor} 
\usepackage{makecell} 
\usepackage{longtable} 
\usetikzlibrary{positioning} 

\usepackage{adjustbox} 
\graphicspath{{./img/}}

\usepackage[skip=5pt plus1pt, indent=10pt]{parskip}
\usepackage{booktabs}
\usepackage{wrapfig}
\usepackage{array}
\usepackage{multirow}
\usepackage[normalem]{ulem}
\useunder{\uline}{\ul}{}
\usepackage{rotating}
\usepackage{algorithm}
\usepackage{algpseudocode}

\usepackage{eurosym}
\usepackage{comment} 

\usepackage{amssymb}
\usepackage{upgreek}
\usepackage{svg}
\usepackage{booktabs}
\usepackage{bigstrut}
\usepackage{colortbl}
\usepackage{pgfplots}
\pgfplotsset{compat=1.18}  

\usepackage{forest}

\title{AI for Business} 
\subtitle{From data to decisions}
\date{\today}
\author{Wouter Verbeke \\ Bart Baesens \\ Johannes De Smedt \\ Jochen De Weerdt \\ Hans Weytjens (chapter author)}

\begin{document}
\maketitle

\chapter{Uncertainty}\label{ch29}
\section{Chapter Objectives}
In this chapter, you learn to
\begin{itemize}
\item identify and distinguish between different types of uncertainty.
\item quantify uncertainty in prediction models, including linear regression, random forests, and neural networks.
\item apply conformal prediction to generate predictions within preset confidence intervals.
\item leverage uncertainty quantification for improved business decision-making.

\end{itemize}

\section{Introduction}
Despite their remarkable successes and ubiquitous use, machine learning models face serious challenges. Level 5 autonomous vehicles encounter difficulties with cameras hindered by inadequate lighting and blockages, and with radars that exhibit low resolution and interference problems. LIDAR sensors, which use laser pulses to create high-resolution 3D maps of the environment, could alleviate some of these issues but remain both expensive and not yet widely available \citep{chougule2023comprehensive}. While these hardware limitations pertain to the sensors themselves, they directly impact the accuracy of the vehicles' machine learning models and have contributed to several widely reported fatal accidents.

In a study focusing on the detection and prognosis of COVID-19 from chest radiographs and chest computed tomography images, \cite{roberts2021common} concluded that none of the models identified in 2,212 studies were of potential clinical use. Among various proposed solutions to improve the COVID-19 models, the authors recommended using confidence intervals to quantify the models' prediction uncertainty, given the often small datasets on which the models were trained. They also urged researchers to recognize that a negative RT–PCR\footnote{Reverse Transcription Polymerase Chain Reaction} test does not necessarily mean that a patient does not have COVID-19. 

Both examples demonstrate the fallibility of machine learning models, including highly accurate and acclaimed ones. The fundamental problem is that most machine learning models make predictions without being aware of the uncertainty associated with those predictions. By design, machine learning models always make a prediction, even in cases that differ significantly from anything seen in the training data or in cases where noisy training data makes predictions unreliable at best. Consequently, they lack a reject option \citep{rejectopion2024} that would allow them to abstain from making a prediction when confidence is insufficient.

The business world is not immune to costly machine learning mishaps either. Think of supply chain planning solutions underestimating demand resulting in out-of-stock situations, fraud (e.g., anti-money laundering) detection systems wrongly flagging legitimate transactions, or credit-scoring algorithms (e.g., for loan approval systems) wrongly denying loans to qualified applicants.

The inherent uncertainty in machine learning models' predictions poses a significant challenge to their broader adoption in business. Despite their impressive accuracy and predictive prowess, these models often overlook the unpredictability and variability of real-world scenarios, resulting in overconfidence. The inability of models to quantify their uncertainty can lead to substantial errors. Consequently, businesses may be reticent to fully incorporate these models into their operations. A potential solution is to develop machine learning models that not only provide predictions but also a measure of the uncertainty associated with these predictions. This would enhance the utility and applicability of machine learning models and allow businesses to make more informed decisions by taking uncertainty into account. Therefore, addressing prediction uncertainty is critical for the future advancement and wider acceptance of machine learning in business.

Before discussing approaches for uncertainty quantification, we will first develop an understanding of the two main types of uncertainty.

\section{Two types of uncertainty}

\cite{hora1996aleatory} distinguished between epistemic and aleatoric uncertainty, a distinction that is still commonly used in machine learning \citep{hullermeier2021aleatoric}. Simply stated, epistemic, or model, uncertainty is caused by a lack of data, whereas aleatoric uncertainty is caused by noisy data. Imagine a model trained on thousands of correctly labeled cat and dog images. It will confidently classify an unseen image of, say, a camel as either a cat or dog, without any indication of uncertainty about this classification. This is an example of \textbf{epistemic uncertainty}: during its training phase, the model did not have access to camel images. If the training set contains incorrectly labeled images (e.g., due to faulty sensors or human labeling errors), the model will likely produce erroneous predictions as well, again without recognizing its errors. This is an example of \textbf{aleatoric uncertainty}.

\subsection{Epistemic uncertainty}
\begin{figure}[htbp]
    \centering
    \includegraphics[width=0.9\textwidth]{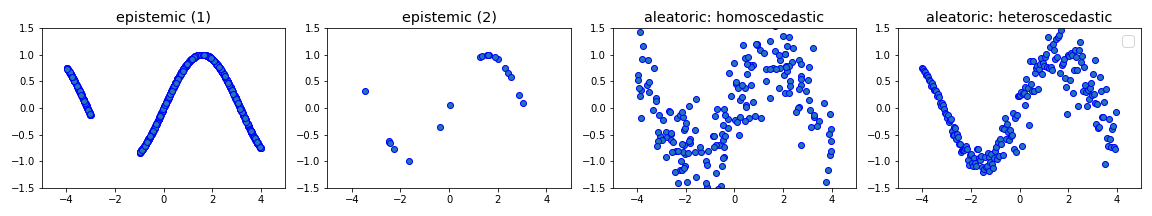}
    \caption{Examples of epistemic and aleatoric uncertainty for a regression task.}
    \label{fig:ch29:uncertaintytypes}
\end{figure}

Epistemic\footnote{The word "epistemic" originates from the Greek word "epistēmē", which means knowledge or understanding. Epistemology is a branch of philosophy concerned with the nature and scope of knowledge.} uncertainty stems from a lack of data as visualized in the data examples in the two left panels in \autoref{fig:ch29:uncertaintytypes}. A model trained on such data is likely to make unreliable predictions for those parts of the domain\footnote{In this context, the "domain" refers to the entire space of possible inputs or conditions (often visualized as the $X$-values or features) that the model is designed to handle. It represents the full range of what the model should be able to predict for, encompassing all potential combinations of characteristics or circumstances relevant to the problem.} where training observations are missing. This is why epistemic uncertainty is also called model uncertainty: the parameters of the model are uncertain, and this uncertainty largely manifests as unreliable or uncertain predictions, especially for parts of the domain where training observations are sparse or missing.

Note that high epistemic uncertainty does not necessarily lead to wrong predictions. If the ground truth behind the examples in \autoref{fig:ch29:uncertaintytypes} is indeed a sine function, some models may correctly interpolate between known observations. However, this is dangerous, as the correct ground truth cannot be determined from the available data. It could be a sine function—but it doesn't have to be. Extrapolating, i.e., making predictions beyond the range of training data, is even riskier.

Epistemic uncertainty is considered \textit{reducible} because it can be reduced by adding more data.

\begin{figure}[htbp]
    \centering
    \includegraphics[width=0.5\textwidth]{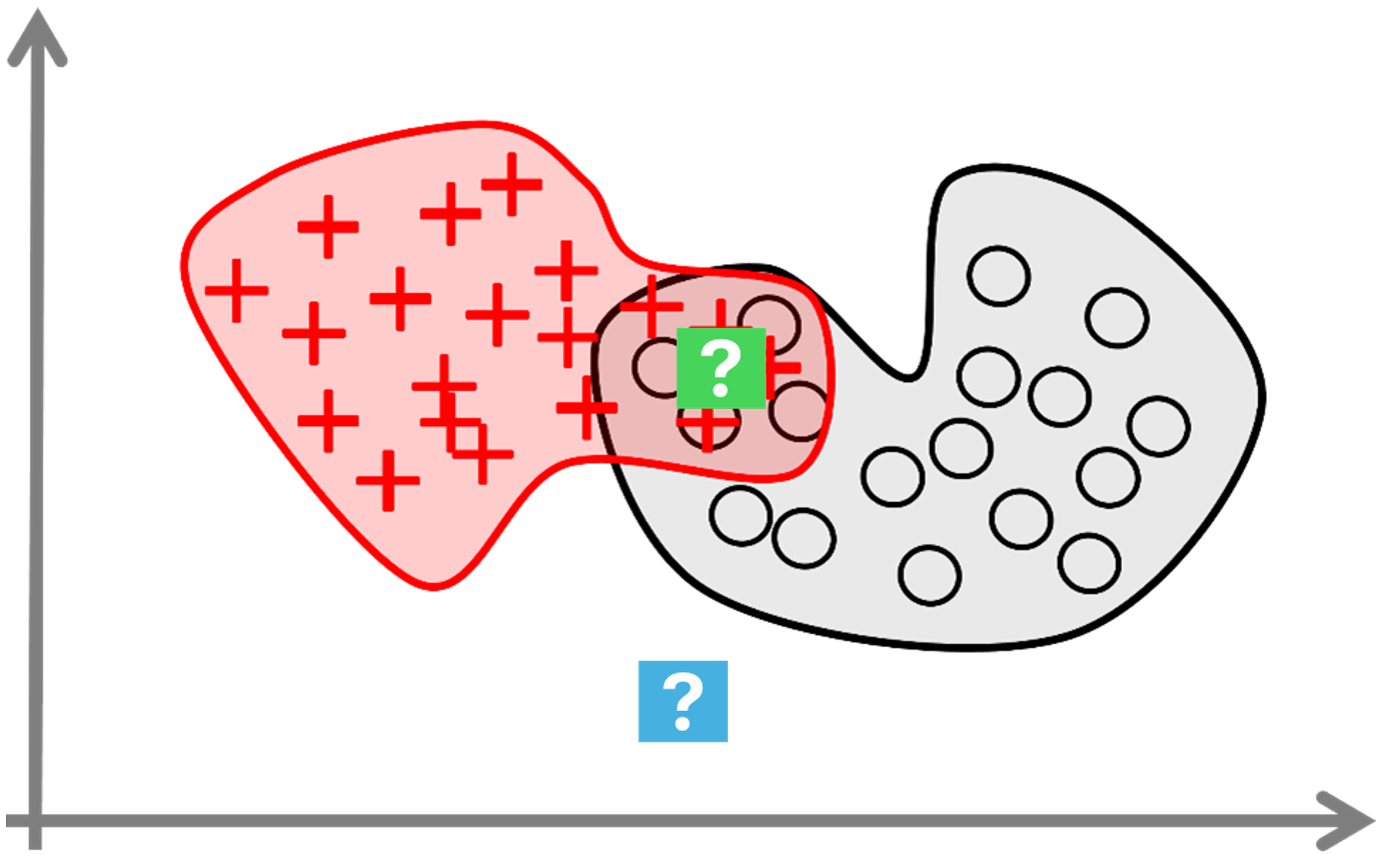}
    \caption{Examples of epistemic and aleatoric uncertainty for a classification task. Image credit: \cite{hullermeier2021aleatoric}}
    \label{fig:ch29:uncertaintytypes_class}
\end{figure}

Epistemic uncertainty is not restricted to regression tasks as in \autoref{fig:ch29:uncertaintytypes}. Classification tasks are susceptible to it as well, as demonstrated in \autoref{fig:ch29:uncertaintytypes_class}. Even though the model performs well in capturing both classes, any prediction for the blue observation (shown as a square) is plagued by high epistemic uncertainty.

The authors of the COVID-19 meta-study that was mentioned in the introduction of this chapter, \cite{roberts2021common}, clearly refer to epistemic uncertainty when they criticize “the prediction uncertainty given the often small datasets”. Models of autonomous vehicles confronted with “inadequate lighting and blockages” are also confronted with epistemic uncertainty, as these situations may differ from what they were trained on. This happened in the 2016 collision of a Tesla car with a truck. In its statement (The Tesla Team, 2016), Tesla referred to “extremely rare circumstances”, which we can interpret as another case of high epistemic uncertainty.

Returning to our examples in business, epistemic uncertainty can lead to underestimated demand if the model, trained primarily on established products, fails to generalize to a radically new product. It can cause fraud detection systems to wrongly flag legitimate transactions as suspicious if the model's training data lacked sufficient examples of certain legitimate but uncommon transaction patterns, thus failing to learn to recognize these as legitimate. A credit scoring system may incorrectly result in rejecting loan applications because the underlying model architecture is too simplistic to adequately capture the complex, non-linear relationships present in the available applicant data.

Epistemic uncertainty can contribute to unfair model behavior when certain groups are underrepresented in the data. In medical research, for instance, men have been enrolled far more often than women \citep{merone2022sex}. The resulting lack of female data increases epistemic uncertainty about female physiology and treatment response. Downstream models or guidelines built on these data may therefore perform worse for women.

\subsection{Aleatoric uncertainty}
Noise or randomness in the data is the root cause of aleatoric\footnote{The term "aleatoric" comes from the Latin word "alea," which means dice or chance.} uncertainty. A coin flip is an archetypal example. The two right-hand images in \autoref{fig:ch29:uncertaintytypes} visualize this notion of randomness: the outcomes for the same or highly similar (same x-value) observations are highly variable. When randomness is equally distributed, we speak of homoscedasticity: every observation across the domain is affected by the same level of randomness. Homoscedasticity need not be the case; different medical facilities with different patient (e.g., age) profiles may use different tests to recognize certain diseases, each of which has a different level of randomness. This is an example of heteroscedasticity where the randomness varies across the domain.

\autoref{fig:ch29:uncertaintytypes_class} shows aleatoric uncertainty for a classification task. In the middle of the image, there is a strong overlap between the classes, creating an area of high aleatoric uncertainty. Consequently, any prediction for the green observation will be highly uncertain. In contrast to epistemic uncertainty, aleatoric uncertainty is irreducible. Gathering more data (e.g., by flipping the coin more often) will not reduce it. In sensor applications, for example, aleatoric uncertainty can arise due to factors such as noise, variability in environmental conditions, or limitations in sensor accuracy. The only way to address aleatoric uncertainty is to combat these factors—for example, by installing more accurate and robust sensors—rather than collecting more readings from the existing sensors. Aleatoric uncertainty appears in our COVID research example in the form of inaccuracies in (or possibly misinterpretation of) RT–PCR tests \citep{roberts2021common}. The low resolution and interference problems of radars in autonomous vehicles \citep{chougule2023comprehensive} also represent a source of heteroscedastic aleatoric uncertainty.

Revisiting the business examples, inventory optimization systems fail when training datasets suffer from inconsistent promotional period labeling or lack real-time competitor pricing data, creating irreducible uncertainty. Similarly, credit scoring models encounter aleatoric uncertainty through noisy income reporting by applicants and missing local economic stress indicators. Errors in loan approval decisions could stem from datasets that exclude alternative payment histories like rent and utilities, leaving inherent information gaps explaining applicants' creditworthiness that no amount of additional training can resolve.

Although aleatoric uncertainty is generally considered irreducible, it can sometimes be mitigated by introducing additional relevant variables into the model. What initially appears to be random noise may stem from systematic influences that are simply unobserved. For instance, in a medical diagnosis setting, test result variability might seem aleatoric, but if patient-specific factors like age, comorbidities, or time since infection are included, the model may better explain the observed variability. Similarly, in the context of autonomous vehicles, environmental conditions like fog density or road surface type—if not previously modeled—could reduce what seemed to be irreducible uncertainty in radar signal interpretation.

\subsection{Estimating uncertainty}
You are now equipped with an understanding of the two kinds of uncertainty. You can recognize them and make concerted efforts to mitigate them. And if mitigation is not feasible, you will have a healthy distrust of your models. You will not deploy your model in France if it was trained on data from Germany. You will be suspicious of a model with unreliable sensor readings or when humans are responsible for data labeling (mistakes are human, after all).

In practice, the total uncertainty usually matters most when making decisions. An autonomous vehicle should transfer control back to the driver whenever it experiences uncertainty, regardless of whether an unexpected object appears on the road (epistemic uncertainty) or it suddenly starts snowing (aleatoric uncertainty).

It is worth bearing in mind that aleatoric and epistemic uncertainty are not fully distinguishable, but rather context-dependent. Consider the roll of a die. This is commonly seen (as we did above) as an example of aleatoric uncertainty. No matter how many times you roll the dice, you can never predict with certainty what number will come up next. However, the uncertainty could be seen as epistemic. If you had perfect knowledge of all the relevant variables (like the exact force and angle of the throw, the weight distribution of the die, the friction between the die and the table, air resistance, etc.), you could theoretically predict the outcome of the roll with certainty. In practice, these variables (and possibly the model) are missing, and the outcome of the dice roll appears random.

As a motivating example for the remainder of this chapter, \autoref{table:ch29:kendallimages} \citep{kendall2017uncertainties}\footnote{Alex Kendall decided to pursue the matter outside of academia and moved on to become the co-founder and CEO of Wayve, a company pioneering AI technology to power self-driving vehicles.} shows uncertainty estimates for an image segmentation task for road scene understanding. Three input images are shown in the first column, with the corresponding ground truth (optimal solution, target for the model) next to it. The middle column shows the model's predictions. The aleatoric uncertainty in column four is high for far-away objects and borders between objects. The model struggles with the sidewalk in the third image. The epistemic uncertainty could be attributed to the weak resemblance to sidewalks in the training dataset. There is also significant aleatoric uncertainty in the same area. As stated before, however, epistemic and aleatoric uncertainty cannot always be separated. The total uncertainty matters and points to the lower quality of the predicted segmentation in the area.

\begin{table}[htp]
\centering
\begin{tabular}{cccccc}
Input  & Ground  & Semantic  & Aleatoric  & Epistemic  \\ 
 image &  truth &  Segmentation &  uncertainty &  uncertainty \\ 
\includegraphics[width=0.185\textwidth]{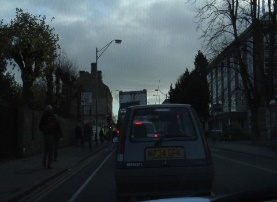}\hspace{-7pt} & 
\includegraphics[width=0.185\textwidth]{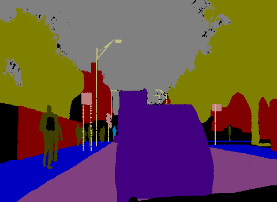}\hspace{-7pt} & 
\includegraphics[width=0.185\textwidth]{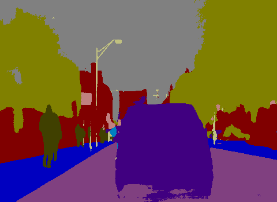}\hspace{-7pt} & 
\includegraphics[width=0.185\textwidth]{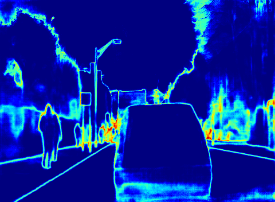}\hspace{-7pt} & 
\includegraphics[width=0.185\textwidth]{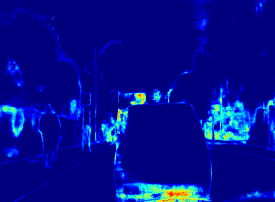} \\

\includegraphics[width=0.185\textwidth]{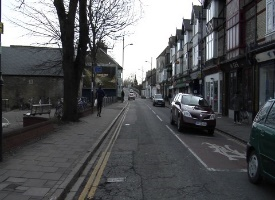}\hspace{-7pt} & 
\includegraphics[width=0.185\textwidth]{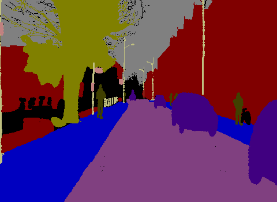}\hspace{-7pt} & 
\includegraphics[width=0.185\textwidth]{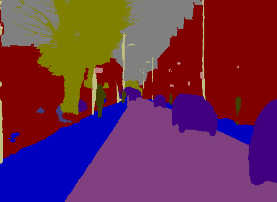}\hspace{-7pt} & 
\includegraphics[width=0.185\textwidth]{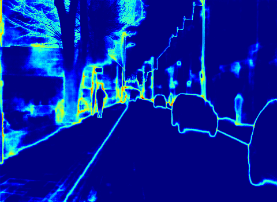}\hspace{-7pt} & 
\includegraphics[width=0.185\textwidth]{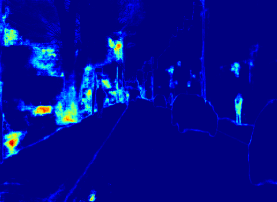} \\

\includegraphics[width=0.185\textwidth]{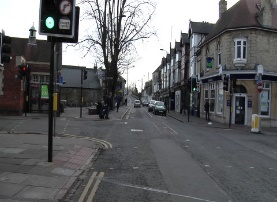}\hspace{-7pt} & 
\includegraphics[width=0.185\textwidth]{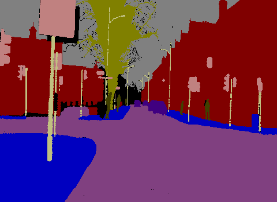}\hspace{-7pt} & 
\includegraphics[width=0.185\textwidth]{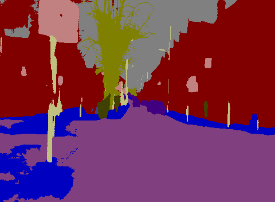}\hspace{-7pt} & 
\includegraphics[width=0.185\textwidth]{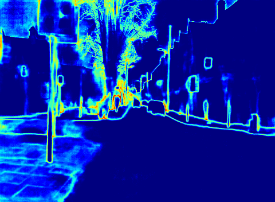}\hspace{-7pt} & 
\includegraphics[width=0.185\textwidth]{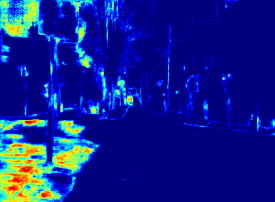} \\
\end{tabular}
\caption{Illustration of aleatoric and epistemic uncertainty for an image segmentation task. Image credit: \cite{kendall2017uncertainties}}
\label{table:ch29:kendallimages}
\end{table}

In the following sections, we will explore uncertainty quantification techniques. We start with the relatively straightforward case of linear regression, before moving on to random forests and, finally, neural networks. The Python code for these examples can be found online\footnote{\url{https://github.com/hansweytjens/AI4B_ch29_uncertainty/}}.

\section{Linear Regression}\label{sec:ch29:OLS}
The beauty of linear regression lies in the existence of analytical, closed-form solutions such as the ordinary least squares (OLS) estimator. Recall that we assume the following linear model:

$$y = X\beta+\varepsilon$$

with $X$ the independent variables (also called regressors or explanatory variables) organized in matrix form, $y$ the dependent variable (also called output or target variable), and $\varepsilon$ the random error term. The OLS estimator for $\beta$, denoted as $b$, minimizes the sum of the squared errors and is given by the formula:

$$b = (X^\top X)^{-1}Xy$$

Under appropriate assumptions (see box below), we can now make predictions for the target variable $y$ for any given observation $x_h= [1, x_{h,1},x_{h,2}, \ldots, x_{h,k}]^\top$ in which $k$ is the number of independent variables:

$$\widehat{y_h} = x_hb$$

\kulbox{\textbf{DRILL-DOWN: The assumptions underpinning Ordinary Least Squares (OLS)} \citep{frost2019regression}\newline 
\begin{itemize}
\item   Linearity, as expressed by the formula $y = X\beta+\varepsilon$.
\item 	The mean of the error term $\varepsilon$ is zero. Otherwise, the model is biased.
\item 	The error term $\varepsilon$ is uncorrelated with the independent variables $X$. Otherwise, $\varepsilon$ has explanatory power which should be captured by $X$ for OLS to work.
\item Individual observations of the error term $\varepsilon$ are uncorrelated. Unexpectedly high sales today (i.e., not explained by the independent variables) should not increase the likelihood of unexpected sales tomorrow.
\item Homoscedasticity: the error term $\varepsilon$ has constant variance.
\item The independent variables are not perfectly correlated with (a combination of) other independent variables. If they were, OLS would be unable to estimate their separate effects.
\item 	The error term $\varepsilon$ is normally distributed. This assumption is required to calculate confidence intervals.
\end{itemize}}

\subsection{Calculating a confidence interval}
Suppose we wish to build a 90\% confidence interval around $\widehat{y_h}$ which corresponds to the expectation that the true, unknown $y_h$ falls within the boundaries of the confidence interval around $\widehat{y_h}$ with 90\% certainty. Suppose further we have $n=1000$ observations in our dataset and $k=5$ independent variables.

The confidence interval around that prediction $\widehat{y_h}$ is given by:

$$\widehat{y_h} = \pm t_{(1-\frac{\alpha}{2},\:n-k-)1}\sqrt{\sigma(1+x^\top_h(X^\top X)^{-1}x_h}$$

The formula requires some explanation: We define $\alpha$ as the significance level; in our example, $\alpha = 10\% = 0.1$. We can now look up $t_{(.95,\:994)} \approx t_{(.95,:\infty)} = 1.645$ in a $t$-distribution table in which $t_{(1-\frac{\alpha}{2},\:n-k-1)}$ is the $1-\frac{\alpha}{2}$ quantile of the $t$ distribution with $n-k-1$ degrees of freedom. $t$-distribution tables are readily available on the internet \citep{devineni1999t} or can be computed using a simple function call available in statistical libraries of most modern programming languages (e.g., scipy.stats.t for Python).

The assumptions outlined in the drill-down box stipulate that $\varepsilon \sim \mathcal{N}(0, \sigma^2 I)$, i.e., the errors are drawn from a normal distribution with mean $0$ and variance $\sigma^2$. Unfortunately, we do not know $\sigma$, and thus we resort to the residual values:

$$e = y - Xb$$

to compute an unbiased estimator for $\sigma^2$ \citep{johnston1984econometric}:

$$s^2 = \frac{e^\top e}{n-k}$$

Substituting $s$ for $\sigma$, we can now use the formula to build confidence intervals in practice.

\subsection{Working with confidence intervals}
In this section, we explore how to construct and interpret confidence intervals for linear regression predictions.

\begin{figure}%
    \centering
    \subfloat[\centering Base case: 90\% confidence interval]{{\includegraphics[width=0.4\textwidth]{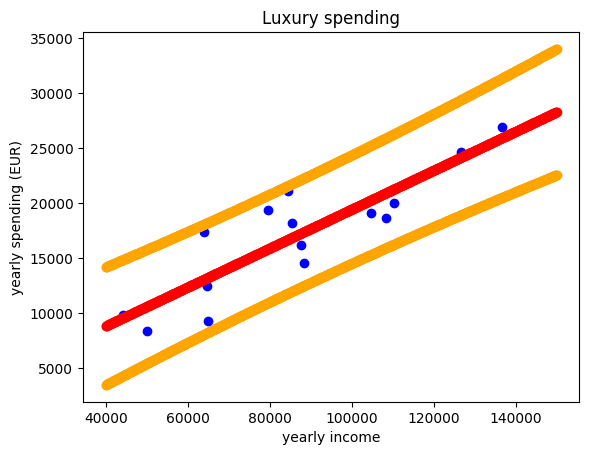} \label{fig:ch29:OLS_basecase_a}}}%
    \qquad
    \subfloat[\centering Lower confidence level: 80\%]{{\includegraphics[width=0.4\textwidth]{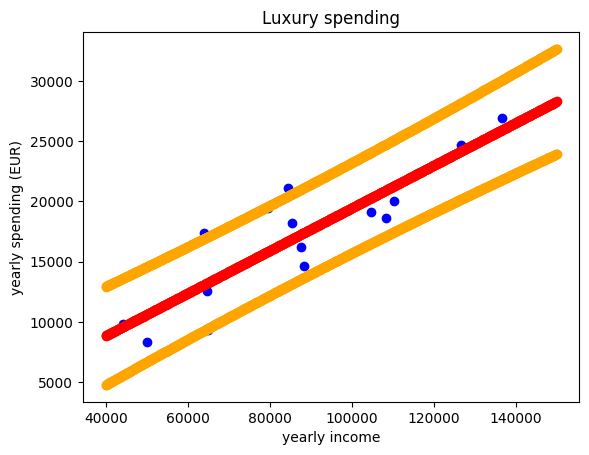} \label{fig:ch29:OLS_basecase_b}}}%
    \caption{Confidence intervals: base case and lower confidence level}%
    \label{fig:ch29:OLS_basecase}
\end{figure}

We wish to estimate a person's annual spending on luxury goods based on the knowledge of that person's yearly income. We have a small set of 15 (fictional) observations for people whose yearly income and luxury goods spending are known. These 15 observations are represented by the blue dots in the two panels in \autoref{fig:ch29:OLS_basecase}. The red line shows the OLS estimates ($\widehat{y_h}$) for luxury goods spending at different income levels. When constructing a 90\% confidence interval, we arrive at the yellow lines in panel\autoref{fig:ch29:OLS_basecase_a}. We can observe that the confidence interval widens marginally when moving away from the central section, which contains the highest density of observations. This corresponds to our intuition that higher uncertainty is found where fewer observations are available.

Panel\autoref{fig:ch29:OLS_basecase_b} demonstrates the effect of changing the confidence level to 80\%. Allowing more observations to be outside of the boundaries narrows the confidence interval. We can see that $3$ out of $15$ observations are located outside the confidence interval.

\begin{figure}%
    \centering
    \subfloat[\centering Fewer observations: epistemic uncertainty]{{\includegraphics[width=0.4\textwidth]{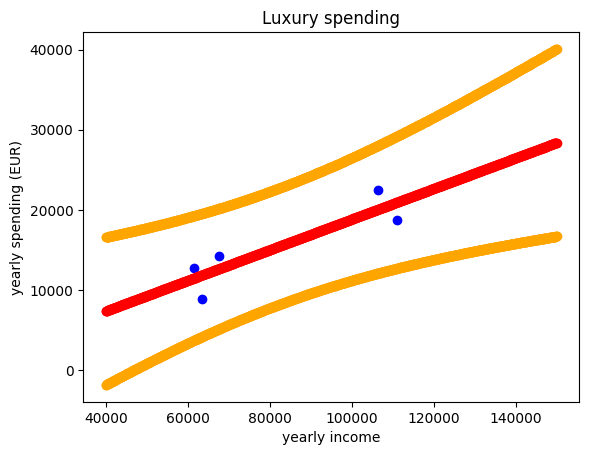} \label{fig:ch29:OLS_unc_a}}}%
    \qquad
    \subfloat[\centering More noise: aleatoric uncertainty]{{\includegraphics[width=0.4\textwidth]{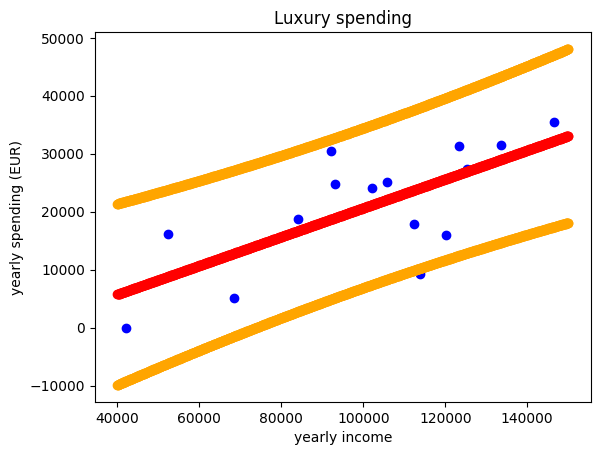} \label{fig:ch29:OLS_unc_b}}}%
    \caption{Confidence intervals: adding epistemic and aleatoric uncertainty}%
    \label{fig:ch29:OLS_unc}
\end{figure}

Confidence intervals do not distinguish between epistemic and aleatoric uncertainty. Nevertheless, both types of uncertainty will manifest in the confidence levels. In panel\autoref{fig:ch29:OLS_unc_a}, we increased the epistemic uncertainty by dramatically reducing the number of observations in the dataset. As a result, the confidence interval becomes wider, increasingly so when moving away from the central part of the domain that contains the observations. A similar effect can be observed in panel\autoref{fig:ch29:OLS_unc_b} where more noise, characteristic of aleatoric uncertainty, was added.

\begin{figure}%
    \centering
    \subfloat[\centering Heteroscedasticity]{{\includegraphics[width=0.4\textwidth]{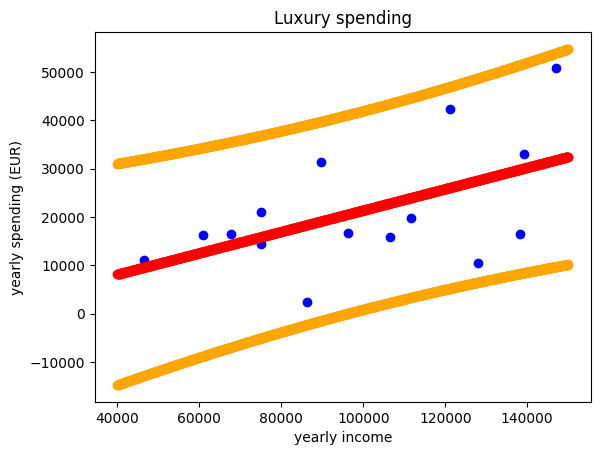} \label{fig:ch29:heterosc_a}}}%
    \qquad
    \subfloat[\centering Extrapolation]{{\includegraphics[width=0.4\textwidth]{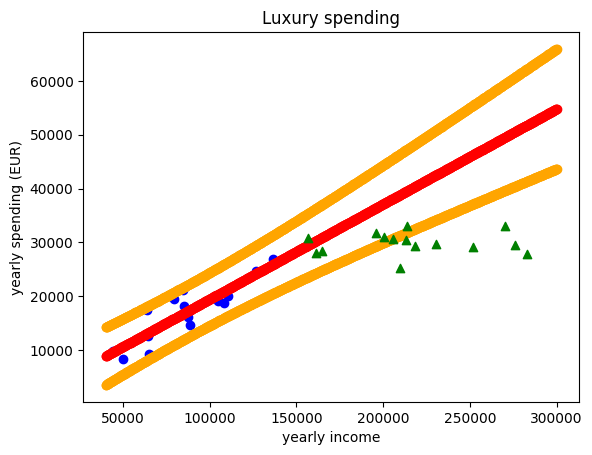} \label{fig:ch29:heterosc_b}}}%
    \caption{: Limitations of confidence intervals: heteroscedasticity and extrapolation.}%
    \label{fig:ch29:heterosc}
\end{figure}

\autoref{fig:ch29:heterosc} provides insight into some of the limitations of linear regression confidence intervals. In panel\autoref{fig:ch29:heterosc_a}, the assumption of homoscedasticity was violated: there is more noise in the targets for the higher yearly income levels. The calculations of the confidence levels, however, assume homoscedasticity, or equal variance for the error term $\varepsilon$. The resulting confidence intervals no longer reflect the uncertainty: they are too wide for the lower yearly incomes. 
When interpreting confidence intervals, care must be taken when considering observations far outside the domain, as the green triangles in panel\autoref{fig:ch29:heterosc_b} can demonstrate. In our (hypothetical) example, luxury consumption flattens out once a yearly income level of around \euro$150k$  is reached. Since there are no observations of people with yearly incomes over \euro$150k$ in the training data, this phenomenon cannot be captured, neither by the point estimates nor the confidence intervals. This is also the reason why linearity is one of the assumptions for OLS. Of course, no machine learning method would be able to predict the values represented by the green triangles.

\subsection{Conclusions}
In situations where the (relatively strong) assumptions for OLS hold, the method proves easy to use, and confidence intervals are a useful tool to deal with uncertainty. When these assumptions are violated, OLS will produce less reliable estimates and confidence intervals. The confidence intervals are not capable of signaling the violation of the assumptions. Therefore, it is important to check the assumptions before making OLS estimates and creating confidence intervals. Checking whether the residuals are truly random and centered around a mean of zero is an important indication of the validity of the OLS regression.

\section{Random Forests}\label{sec:ch29:rf}

\subsection{Random forests}

A random forest is an ensemble method that combines multiple decision trees to create a more accurate and robust predictive model. Random forests employ bagging (bootstrap aggregating), which involves training each tree on a different bootstrap sample drawn from the original dataset. What distinguishes random forests from simple bagging is the addition of feature randomness: at each node split, only a random subset of features is considered as potential splitting variables. This dual randomness—in both training samples and feature selection—increases diversity among the individual trees. This approach reduces overfitting and enhances model stability and robustness, typically improving predictive accuracy compared to single decision trees. For final predictions, the ensemble aggregates results from all trees through majority voting in classification tasks or by averaging predictions in regression problems.

\kulbox{\textbf{FUN FACT: Bootstrapping} The term ``bootstrapping'' in machine learning—and more broadly in statistics and computing—traces back to the 19th-century idiom “to pull oneself up by one’s bootstraps,” originally coined as a tongue-in-cheek description of an impossible feat: how could you lift yourself off the ground simply by tugging on your own bootstraps? In the context of random forests (and other bagging methods),``bootstrapping'' captures exactly that spirit: we repeatedly draw new training sets—bootstrap samples—from our original data (with replacement) and let each decision tree “pull itself up” into a full-fledged model. By relying solely on the data at hand—no outside labels or assumptions—each tree both stands on its own and, when aggregated, produces a more robust overall predictor.}

We illustrate a random forest's workings using the Iris data set (Fisher, 1988) as an example. The data set consists of 50 observations from each of three species of Iris flowers (Iris Setosa, Iris Virginica, and Iris Versicolor) with measurements of four features (independent variables): the lengths and widths of the sepals and petals. \autoref{tab:ch29:iris} shows a selection of five observations from the dataset.

\begin{table}
\begin{center}
\begin{tabular}{|c|c|c|c|c|c|}
    \hline
    \textbf{Sepal Length} & \textbf{Sepal Width} & \textbf{Petal Length} & \textbf{Petal Width} & \textbf{Target} & \textbf{Target Name} \\
    \textbf{(cm)} & \textbf{(cm)} & \textbf{(cm)} & \textbf{(cm)} &  &  \\
    \hline
    5.1 & 3.5 & 1.4 & 0.2 & 0 & Setosa \\
    4.9 & 3.0 & 1.4 & 0.2 & 0 & Setosa \\
    7.0 & 3.2 & 4.7 & 1.4 & 1 & Versicolor \\
    6.4 & 3.2 & 6.0 & 2.5 & 2 & Versicolor \\
    6.3 & 3.3 & 6.0 & 2.5 & 2 & Virginical\\
    \hline
\end{tabular}
\caption{A selection of five observations of the Iris dataset.}
\label{tab:ch29:iris}
\end{center}
\end{table}

We randomly split the dataset into a training set and test set of 30 and 120 observations\footnote{The very limited training set size induces epistemic uncertainty.} respectively, and train a random forest of 100 trees with a maximal depth of two. We used the RandomForestClassifier from scikit learn \citep{scikit-learn}. 

\begin{figure}[htbp]
    \centering
    \includegraphics[width=0.8\textwidth]{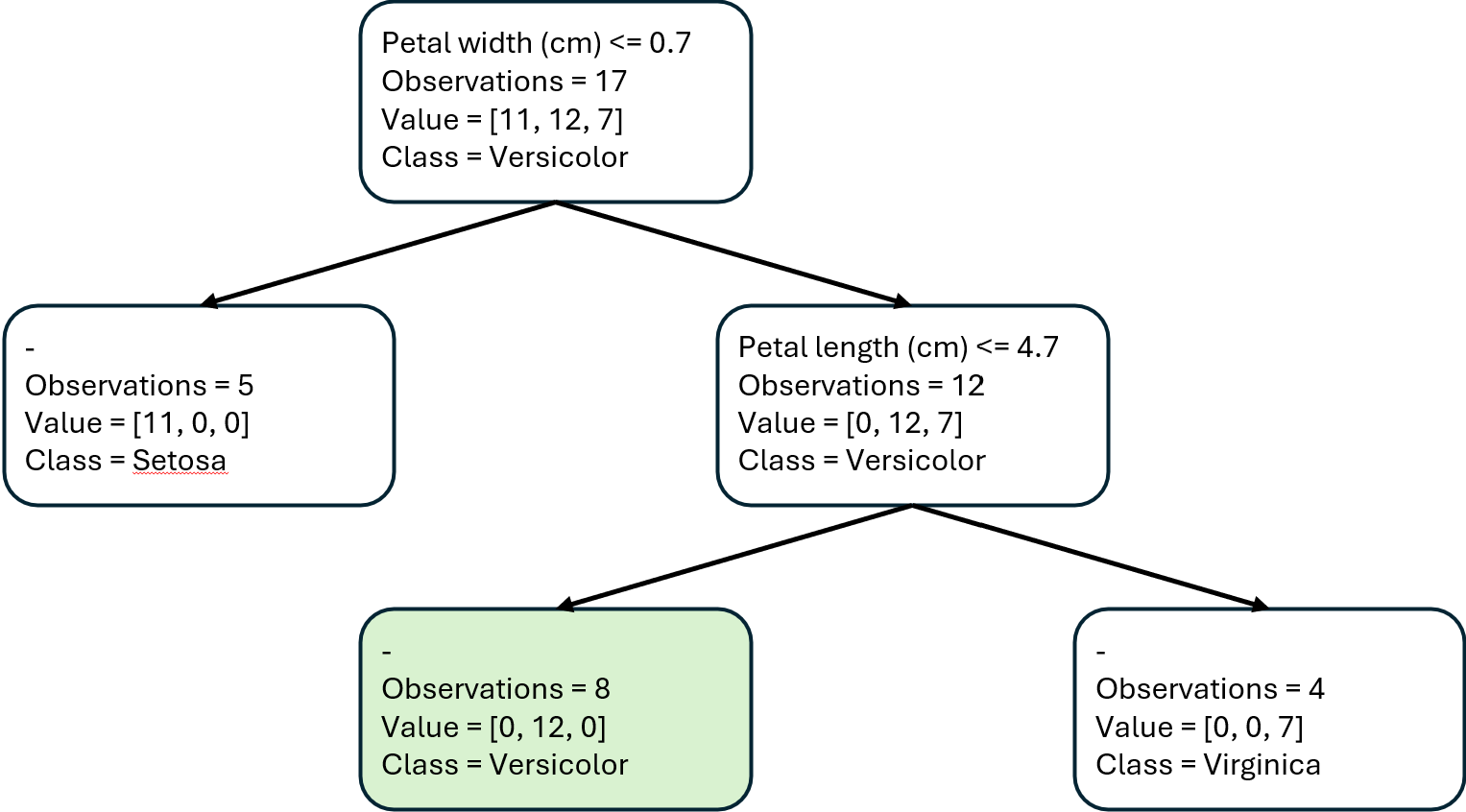}
    \caption{First tree of the random forest. The tree was trained on a subset of 17 observations. 'Value' counts the observations for each class from the original full training dataset that fit the criteria to be at a specific node.}
    \label{fig:ch29:RF_0}
\end{figure}

We want to classify the flower with features displayed in \autoref{tab:ch29:iris_sample}. In order to do so, our random forest will collect the classifications of all 100 constituent trees. The first tree is pictured in \autoref{fig:ch29:RF_0}. Since the petal width of our flower is 1.3 cm, which is larger than 0.7, we move to the right branch. From there, we move left as our petal length of 4.1 cm is smaller than 4.7. We end up in a leaf with 12 observations from the complete training set, all of which carry label 1 (Versicolor). Note that the total number of observations from the bootstrap sample for this particular tree is eight.

\begin{table}
\begin{center}
\begin{tabular}{|c|c|c|c|}
    \hline
    \textbf{Sepal Length} & \textbf{Sepal Width} & \textbf{Petal Length} & \textbf{Petal Width} \\
    \textbf{ (cm)} & \textbf{ (cm)} & \textbf{(cm)} & \textbf{(cm)} \\
    \hline
    5.6 & 3.0 & 4.1 & 1.3 \\
    \hline
\end{tabular}
\caption{Test observation.}
\label{tab:ch29:iris_sample}
\end{center}
\end{table}

\begin{figure}[htbp]
    \centering
    \includegraphics[width=0.8\textwidth]{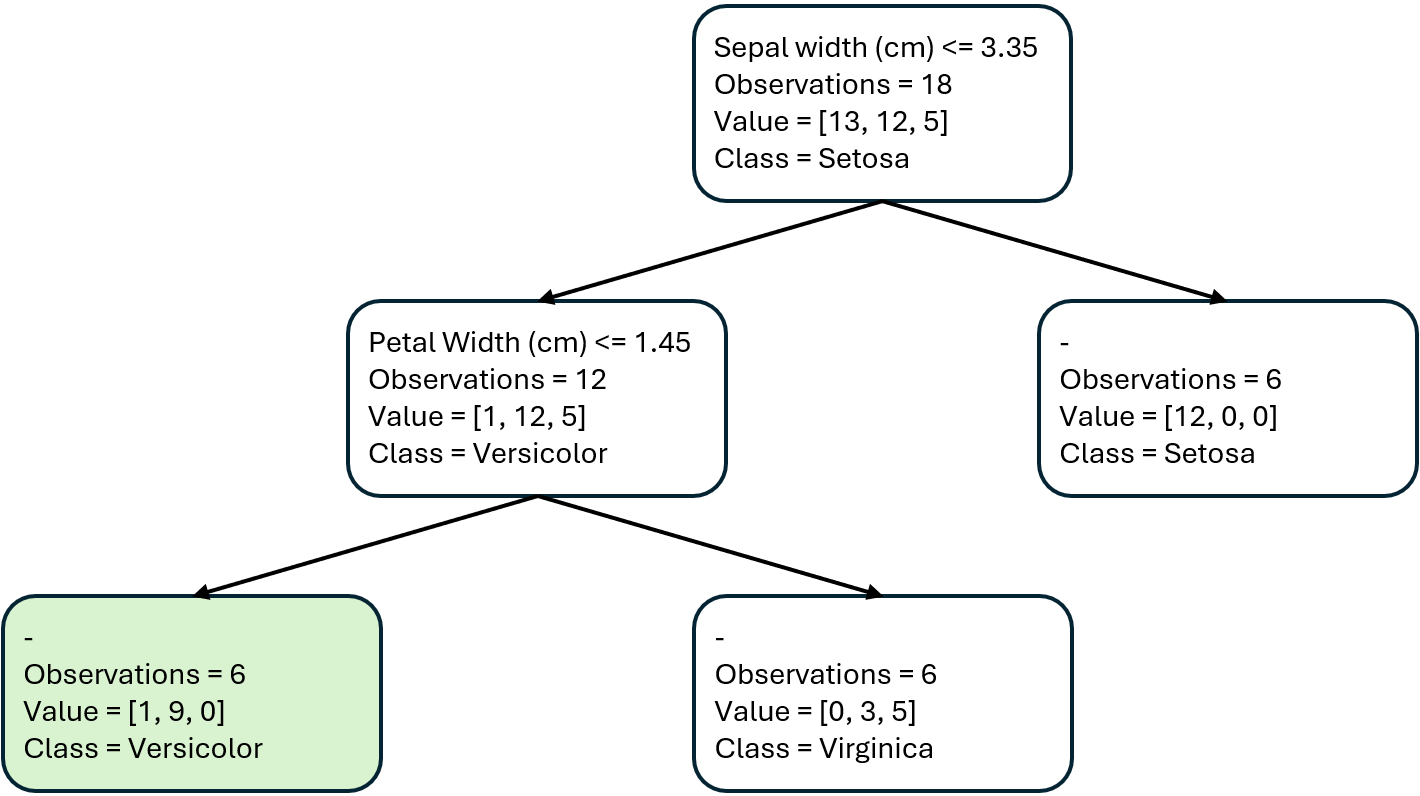}
    \caption{Third tree of the random forest}
    \label{fig:ch29:RF_1}
\end{figure}

Similarly, our flower ends up in the leftmost leaf of the third tree in \autoref{fig:ch29:RF_1}. Based on the observations from the training set found here, the probabilities of being classified as Setosa, Versicolor, or Virginica are represented by the probability vector $[0.1, 0.9, 0.0]$, obtained by dividing the count vector $[1, 9, 0]$ by the sum of the elements, $10$. The prediction of this tree will thus be Versicolor (highest probability class). Note that this leaf demonstrates how a decision tree can be seen as a probabilistic predictor.

Calculating the predictions of all 100 trees and selecting the most frequently predicted class (majority voting) yields the random forest's final prediction: our flower is classified as a Versicolor.

The question that now arises is: how certain is this answer? Is this prediction suffering from epistemic or aleatoric uncertainty?

\subsection{Calculating uncertainty}
For classification tasks, both types of uncertainty, epistemic and aleatoric, can be estimated using entropy concepts from information theory \citep{depeweg2018decomposition}.

The total uncertainty for the prediction of observation $x_h$ can be defined as the (Shannon) entropy $H_h$ of the predicted probability distribution. It is given by the formula \citep{shaker2020aleatoric}:

$$H_h = - \sum\limits_{y}p(y\mid x_h)\text{log}_2 p(y\mid x_h) $$

where $p(y\mid x_h)$ represents the probability of outcome (or class) $y$ for observation $x_h$. \autoref{tab:ch29:entropy_interpretation} \citep{weytjens2022learning} provides intuition about the entropy using a simplified example of an ensemble composed of four models. The formula for $H_h$ was applied to the mean prediction in the second column. 

\begin{table}
\begin{center}
\begin{tabular}{|c|c|c|c|c|c|}
    \hline
    \textbf{Ens-}&\textbf{Probabilities} & \textbf{Mean} & \textbf{$H_h$:} & \textbf{$C_h$:} & \textbf{$I=H-C_h$:} \\
    \textbf{emble}&\textbf{predicted by} & \textbf{prediction} & \textbf{total} & \textbf{conditional} & \textbf{ mutual} \\
    &\textbf{individual models} &  & \textbf{entropy} & \textbf{entropy } & \textbf{information} \\
    & \textbf{in ensemble}&  & \textbf{(total} & \textbf{(aleatoric} & \textbf{(epistemic} \\
    &&  & \textbf{uncertainty)} & \textbf{uncertainty)} & \textbf{uncertainty)} \\
    \hline
    1 & (1, 0), (1, 0), (1, 0), (1, 0) & (1, 0) & 0 (low) & 0 (low) & 0 (low)\\
    2 & (.5, .5), (.5, .5), (.5, .5), (.5, .5) & (.5, .5) & .69 (high) & .69 (high) & 0 (low)\\
    3 & (1, 0), (0, 1), (1, 0), (0,1) & (.5, .5) & .69 (high) & 0 (low) & .69 (high)\\
    \hline
\end{tabular}
\caption{Interpretation of entropy in terms of uncertainty: simplified example.}
\label{tab:ch29:entropy_interpretation}
\end{center}
\end{table}

Inspecting the predicted probabilities in \autoref{tab:ch29:conditional_entropy} helps to understand the formula for the conditional entropy that corresponds to the aleatoric uncertainty:

$$C_h = - \frac{1}{M}\sum\limits_{1}^{M}\sum\limits_{y}p(y\mid h_i,x_h)\text{log}_2 p(y\mid h_i,x_h) $$

with $h_i$ a hypothesis, or model, $i$, with $M$ being the number of models in the ensemble. This formula essentially calculates the mean value of the entropies of the predicted probability vectors of each individual model. It is easy to check the calculations in \autoref{tab:ch29:conditional_entropy}. The four individual models of ensembles 1 and 3 exhibit no aleatoric uncertainty, each constituent model is certain of its prediction. In ensemble 2, however, all models suffer from high aleatoric uncertainty.

Mutual information, another concept from information theory, corresponds to the epistemic uncertainty and is defined as the difference between the total and the conditional entropy:

$$I_h = H_h - C_h$$

The mutual information captures how much the individual models' predictions diverge from each other. The greater the divergence, the higher the epistemic (model) uncertainty.

\begin{table}
\begin{center}
\begin{tabular}{|c|c|c|c|c|}
    \hline
    \textbf{Tree} & \textbf{P(Setosa)} & \textbf{P(Versicolor)} & \textbf{P(Virginica)} & \textbf{Entropy} \\
    \hline
     1 & 0.00 & 1.00 & 0.00 & 0.00\\
     2 & 0.00 & 0.92 & 0.08 & 0.39 \\
     3 & 0.10 & 0.90 & 0.00 & 0.47 \\
     4 & 0.00 & 0.79 & 0.21 & 0.75 \\
     5 & 0.00 & 1.00 & 0.00 & 0.00 \\
     ... & ... & ... & ... & ... \\
    \hline
    \textbf{Mean:} & 0.02 & 0.94 & 0.04 & \textbf{0.19} \\
    \hline
\end{tabular}
\caption{Calculation of $C_h$: the conditional entropy (aleatoric uncertainty) for our test observation.}
\label{tab:ch29:conditional_entropy}
\end{center}
\end{table}

\begin{table}
\begin{center}
\begin{tabular}{|c|c|l|}
    \hline
    \textbf{Entropy} & \textbf{Value} & \textbf{Calculation} \\
    \hline
    \textbf{$H_h$} & 0.38 & Entropy of mean probability vector [0.02, 0.94, 0.04] in \autoref{tab:ch29:conditional_entropy}\\
    \textbf{$C_h$} & 0.19 & Mean of individual entropies in \autoref{tab:ch29:conditional_entropy} \\
    \textbf{$I_h$} & 0.19 & $I_h = H_h - C_h$ \\
    \hline
\end{tabular}
\caption{Calculation of $H_h$, $C_h$, $I_h$.}
\label{tab:ch29:entropy}
\end{center}
\end{table}

We now know how to calculate the (epistemic and aleatoric) uncertainty for any individual observation. The ``Versicolor'' prediction for the flower described in Table 2 has a total uncertainty of 0.38. \autoref{tab:ch29:entropy} and \autoref{tab:ch29:conditional_entropy} show the calculations.

Can we trust the prediction? To answer this question, we need to evaluate the quality of the uncertainty prediction and then calibrate it to make it interpretable and actionable.

\subsection{Accuracy-rejection curves}\label{subsec:accrej}

\begin{figure}[htbp]
    \centering
    \includegraphics[width=\textwidth]{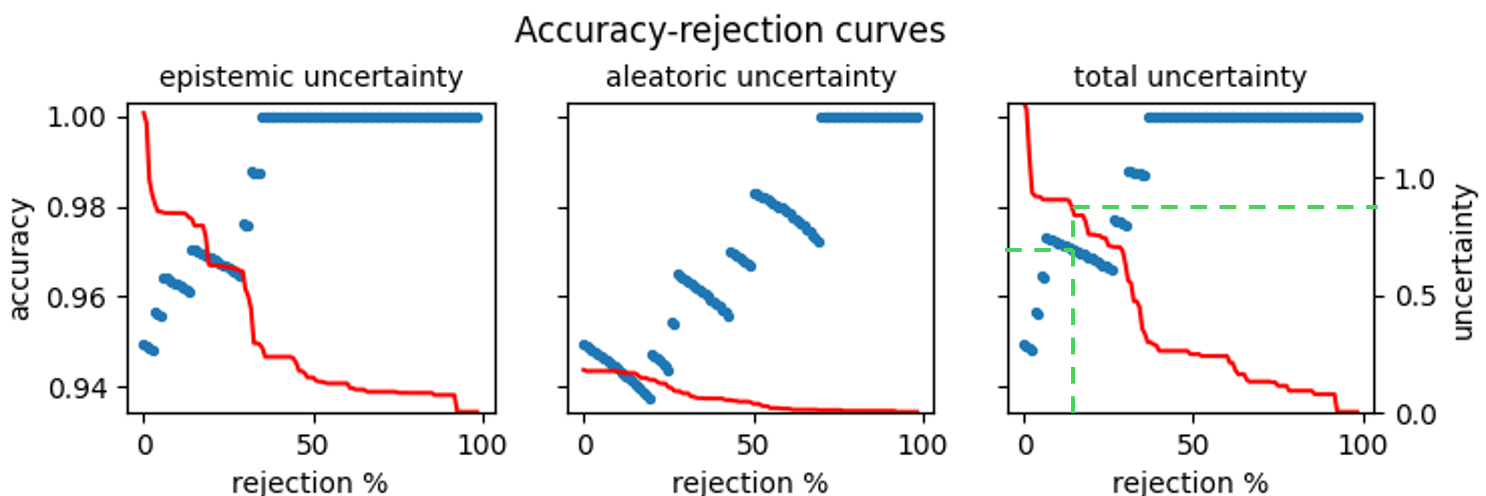}
    \caption{Accuracy-rejection curves for epistemic, aleatoric, and total uncertainty, based on 120 observations in the test set.}
    \label{fig:ch29:acc_rej_RF}
\end{figure}

Because there is no ground truth for uncertainty, uncertainty estimates cannot be evaluated directly. Accuracy-rejection curves \citep{nadeem2009accuracy} offer an indirect method to assess their quality. These curves plot average model accuracy (or another chosen performance metric) on the test set as a function of the proportion of predictions rejected based on uncertainty. Predictions are ranked from most to least uncertain, and the most uncertain are progressively removed. \autoref{fig:ch29:acc_rej_RF} shows these curves for the random forest classifier on a test set of 120 observations. The blue lines show how accuracy improves as more uncertain predictions are rejected; a steeper, more consistently increasing curve suggests more reliable uncertainty estimates. The red lines show the corresponding estimated uncertainty levels for each rejection proportion.

In terms of both steepness and consistent increase, the epistemic uncertainty (left panel) appears to outperform the aleatoric uncertainty (center panel) in \autoref{fig:ch29:acc_rej_RF}. The initial steepness of its blue line reflects that the epistemic uncertainty effectively identifies the least accurate predictions. The relatively smooth and consistent rise in accuracy with increased rejection indicates few 'valleys' where discarding more data leads to lower performance. While aleatoric uncertainty does contribute, its impact is less pronounced. This limited effect is also evident in the total uncertainty (right panel), where incorporating aleatoric uncertainty offers little improvement over using epistemic uncertainty alone.

With the accuracy-rejection curves at our disposal, uncertainties become actionable: we can now relate uncertainty thresholds to expected model performance. For example, suppose an organization requires 97\% prediction accuracy. Referring to the third panel of \autoref{fig:ch29:acc_rej_RF}, this target can be met by rejecting predictions with total uncertainty above 0.9 (as shown on the right-hand y-axis). This threshold corresponds to a rejection rate of about 14\% (horizontal axis), meaning that by discarding the 14\% most uncertain predictions, the remaining predictions are expected to meet the desired 97\% accuracy.

\section{Neural Networks}\label{sec:ch29:nn}

\subsection{Regression}

In this section, we investigate how to extract uncertainty estimates from artificial neural networks (NNs), at least at a conceptual level. The complete mathematical derivations and intricacies are beyond the scope of this book.

\subsubsection{Epistemic uncertainty}

\begin{figure}[htbp]
    \centering
    \includegraphics[width=0.9\textwidth]{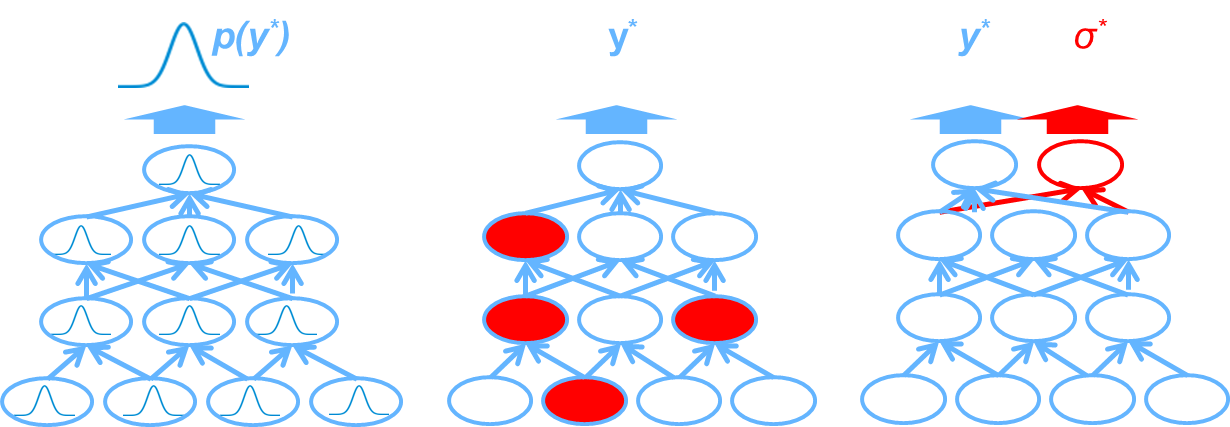}
    \caption{Bayesian neural networks. Left: BNNs model weight uncertainty, yielding predictive distributions.
Center: Dropout deactivates random nodes during training and remains active at inference to approximate uncertainty.
Right: Aleatoric uncertainty is modeled by predicting both the output 
 and its variance.}
    \label{fig:ch29:BNN}
\end{figure}

A classical deterministic NN will always make a prediction, without realizing whether an observation $x_h$ is well within the domain it was trained on, or rather far away from it. In other words, deterministic NNs are ignorant of their model or epistemic uncertainty. To quantify the epistemic uncertainty, we must introduce stochasticity to NNs. We aim to infer a distribution, rather than a single scalar number, for every weight (parameter) in the NN, based on the data in the training set. To compute this distribution, we apply the Bayesian rule and create Bayesian neural networks (BNNs):

$$p(\omega\mid X,y) = \frac{p(y\mid X,\omega).p(\omega)}{p(y,X)} $$

in which $p(\omega\mid X,y)$ stands for the probability distribution of the model weights $\omega$ given the dataset with observations $X$ and corresponding targets $y$. $p(\omega\mid X,y)$ represents the model's ``knowledge''. A narrow and tall peak around the mean value for $\omega$ reflects low uncertainty, whereas a broad, flat distribution indicates high uncertainty. The left panel of \autoref{fig:ch29:BNN} shows how a BNN's model weights are learned as distributions $p(\omega\mid X,y)$ rather than point estimates $\omega$.

Bayesian inference involves marginalizing over the posterior distribution of $\omega$ to obtain a prediction $y_h$ for a given $x_h$

\begin{equation}\label{eq:ch29:bay_inf}
    p(y_h\mid x_h,X,y)=\int\limits_{\omega} p(y_h\mid x_h, \omega).p(\omega\mid X,y).d\omega
\end{equation}

The mean of this non-Gaussian distribution (the outcome of the NN in left panel of \autoref{fig:ch29:BNN}) can be considered as a point estimate, while its variance provides a measure of that estimate’s epistemic uncertainty. Unfortunately, in most practical cases, an analytical solution for $p(y_h\mid x_h,X,y)$ in this equation is intractable. In this form, BNNs cannot be used, we need a clever approximation.

\cite{gal2016dropout} showed mathematically how dropout, in combination with an additional L2 regularizer, can be used to approximate BNNs. Dropout \citep{srivastava2014dropout} is a widespread regularization technique to prevent overfitting in NNs. 

Dropout (center panel of \autoref{fig:ch29:BNN}) is a regularization technique in which, for every batch, a fraction of random nodes (represented here by the red nodes), along with their incoming and outgoing connections, is ignored during training. In a deterministic setting, the dropout mechanism is switched off for inference, yielding point estimates as predictions. In the Bayesian setting, the dropout mechanism remains active during inference. Every forward pass produces a different output, and multiple stochastic passes allow us to approximate the predictive distribution. The right panel shows how aleatoric uncertainty $\sigma_h$, which captures inherent data noise, is estimated. This is achieved by doubling the output layer to predict both $y_h$ and $\sigma_h$ for each input. 

After training a BNN using dropout and L2 regularization, inference is performed using $T$ stochastic forward passes with dropout enabled. The mean of the resulting outputs approximates the expected value of $y_h$ from \autoref{eq:ch29:bay_inf}, providing a robust point estimate:

\begin{equation}\label{eq:ch29:avg}
\mathbb{E}_{p(y_h\mid x_h,X,y)}[y_n]\approx\frac{1}{T}\sum_t\mathcal{H}(x_h,\widehat{\omega})
\end{equation}

in which $\mathcal{H}(x_h,\widehat{\omega})$ represents the result of a forward pass of the NN $\mathcal{H}$ with (randomly dropped-out) weights $\widehat{\omega}$. The variance of the posterior predictive distribution (\autoref{eq:ch29:bay_inf}) can be approximated as:

\begin{equation}\label{eq:ch29:var}
\text{Var}_{p(y_h\mid x_h,X,y)}[y_n]\approx\sigma_h^2+\frac{1}{T}\sum_t\mathcal{H}(x_h,\widehat{\omega})^2-\left( \frac{1}{T}\sum_t\mathcal{H}(x_h,\widehat{\omega})\right)^2
\end{equation}

The last two terms $\frac{1}{T}\sum_t\mathcal{H}(x_h,\widehat{\omega})^2-\left( \frac{1}{T}\sum_t\mathcal{H}(x_h,\widehat{\omega})\right)^2$ represent the sample variance of the $T$ stochastic forward passes and quantify the epistemic or model uncertainty.

\subsubsection{Aleatoric uncertainty}

The $\sigma_h^2$ in \autoref{eq:ch29:var} represents the aleatoric uncertainty for $x_h$. We can learn it by modifying the network architecture to produce an additional output for the uncertainty estimate (right panel of \autoref{fig:ch29:BNN}) and including it in the loss function:

\begin{equation}\label{eq:ch29:loss}
\mathcal{L} =  \min \frac{1}{N} \sum \frac{1}{2\sigma_h^2}\left(y_h-\mathcal{H}(x_h)\right)^2 +\frac{1}{2}\text{log}\,\sigma_h^2 
\end{equation}

\kulbox{\textbf{DRILL-DOWN: The loss function for NN}\newline
In the maximum likelihood estimator (MLE) approach, the loss curve in \autoref{eq:ch29:loss} follows directly from the assumption of a data distribution with Gaussian noise. Most NNs, however, will assume homoscedasticity, which means $\sigma$ is a constant. Under that assumption, the loss curve can be simplified to $\mathcal{L} =  \min \frac{1}{N} \sum\left(y_h-\mathcal{H}(x_h)\right)^2$ which corresponds to the ubiquitous sum of squared errors.
}
\vspace{3mm}

\autoref{eq:ch29:loss} merits closer inspection: for observations $x_h$ in the noisier parts of the domain, the model predicts higher aleatoric uncertainties $\sigma_h^2$, thus reducing the contribution of these observations to the total loss $\mathcal{L}$. This process, known as \textit{loss attenuation}, should result in more accurate predictions while also quantifying the aleatoric uncertainty $\sigma_h^2$. The second term in \autoref{eq:ch29:loss} serves as a regularizer, preventing the model from simply predicting a high $\sigma_h^2$ to minimize $\mathcal{L}$.

This mathematically grounded approach allows us to train BNNs while simultaneously optimizing both accuracy and uncertainty quantification.

\subsubsection{Example}
\begin{figure}[htbp]
    \centering
    \includegraphics[width=\textwidth]{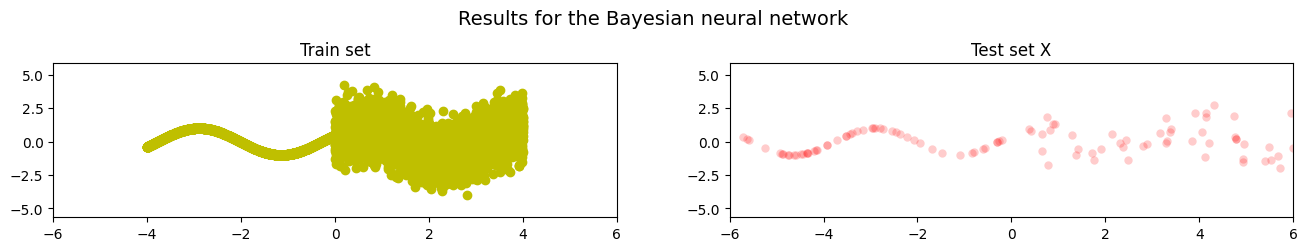}
    \caption{Sinusoidal generative function: training set (left) and test set (right). There are no observations outside of [-4, 4] in the training dataset, creating epistemic uncertainty outside these boundaries. Furthermore, the interval [0, 6] is characterized by high aleatoric uncertainty in both sets.}
    \label{fig:ch29:BNN_0}
\end{figure}

For visualization purposes, we simulate data using by randomly sampling from a univariate sinusoidal function. The training set suffers from both epistemic and (heteroscedastic) aleatoric uncertainty, as shown in \autoref{fig:ch29:BNN_0}: There are no observations outside the domain [-4, 4] and there are high noise levels for observations in [0, 4]. The domain of the test set is wider [-6, 6], and also suffers from noise in the [0, 6] region. We use a simple BNN with loss attenuation containing three fully connected hidden layers with 100 nodes each.

After training the model with dropout, we perform $T = 50$ forward passes for every observation $x_h$ in the test set, and calculate the point estimates by averaging, as prescribed by \autoref{eq:ch29:avg}. 

\begin{figure}[htbp]
    \centering
    \includegraphics[width=\textwidth]{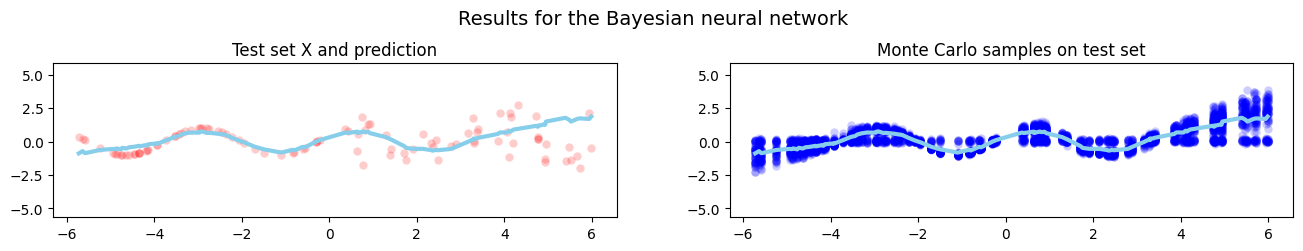}
    \caption{The point estimates $y_h$  for the predictions (left) are the averages of $T$ forward passes for every observation $x_h$ in the test set (right).}
    \label{fig:ch29:BNN_1}
\end{figure}

The resulting blue line is a very good approximator of the generating function in [-4, 4] (left panel in \autoref{fig:ch29:BNN_1}). This is no surprise for [-4, 0], given the lack of uncertainty in that part of the domain. There were enough training examples in the training set and a model with sufficient capacity to arrive at good predictions. At first sight, what is slightly surprising is the good fit in [0, 4], in that part of the domain where the training data was very noisy. However, that noise had a Gaussian distribution around the true generating function. The BNN turns out to average the distribution of these targets, thus approximating the generating function’s curve.

\begin{figure}[htbp]
    \centering
    \includegraphics[width=\textwidth]{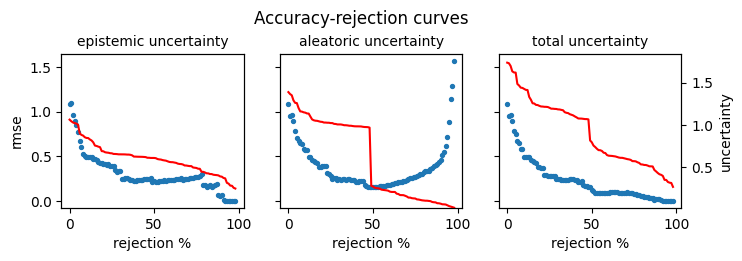}
    \caption{Accuracy-rejection curves for epistemic, aleatoric, and total uncertainty, based on 100 observations in the test set.}
    \label{fig:ch29:acc_rej_BNN}
\end{figure}

According to \autoref{eq:ch29:var}, the sample variance of the $T$ stochastic forward passes quantifies the epistemic, or model, uncertainty. We can see how that variance increases for observations $x_h$ further away from the domain of the training set [-4, 4]. The relatively smooth downward-sloping accuracy-rejection curve for the epistemic uncertainty (left panel of \autoref{fig:ch29:acc_rej_BNN}) confirms the quality of the epistemic uncertainty estimates.

The picture for the aleatoric uncertainty (center panel of \autoref{fig:ch29:acc_rej_BNN}) is less clear-cut. When successively rejecting the observations with the highest aleatoric uncertainty, the overall accuracy of the model increases (lower RMSE). However, after rejecting about 50\% of the observations, the aleatoric uncertainty drops dramatically. This reflects the transition from the [-6, 0] to the [0, 6] range of the domain. At this stage, the aleatoric uncertainty is no longer helpful. This can be explained by the interplay between both kinds of uncertainty. In zone [4, 6], both epistemic and aleatoric uncertainty are high: for lack of observations, there is no way for the model to evaluate the aleatoric uncertainty. That is why, in most cases, it is advisable to first consider the total uncertainty. In our example, too, the total uncertainty delivers the best accuracy-rejection curve.

\subsection{Classification}
In the previous section on regression BNNs, we learned how to approximate the distribution for $y_h$ by performing $T$ forward passes for every observation $x_h$, and then calculating a point estimate by averaging. We can use the same technique for classification tasks. Instead of averaging scalar predictions, we average the probability vectors predicted by the model. The most likely class in the resulting average probability vector determines the final prediction of the BNN.

The attentive reader may have noticed that BNNs are in fact ensembles. By leaving the dropout mechanism active during inference, every forward pass is performed using a different model. The final prediction of the ensemble is the average of the $T$ individual predictions.
The entropy concepts for classification tasks introduced in the previous section on random forests are applicable to all ensemble methods, including BNNs. Calculating entropy to quantify uncertainty is an \textit{ex-post} technique, meaning that it is done after the model has made its prediction. As such, this uncertainty cannot be considered during training to optimize for both accuracy and uncertainty simultaneously. However, further modifications to BNNs do make this simultaneous optimization possible but are beyond the scope of this chapter\footnote{Further reading references: \cite{kendall2017uncertainties}, \cite{dorman2025bayesian}.}. 

\section{Conformal Prediction}
In practical business applications, confidence intervals (regression) or sets (classification) are often used to quantify uncertainty in estimates. For example:

\begin{itemize}
\item Estimating next quarter's sales: "We expect sales to be \euro 1.2 million $\pm$ \euro 100,000 with 95\% confidence."
\item Projecting employee turnover: "We estimate next year's turnover rate to be 12\% $\pm$ 2\% with 90\% confidence."
\item Customer Intent Tagging (a multi-class classification where a set of intents is correct in 90\% of cases):
"For any incoming customer message (email, chat), our AI system is designed to identify the customer's intent. Instead of just a single intent, the system outputs a 90\% confidence prediction set of possible intents. For example:
    \begin{itemize}
    \item For a simple payment issue: "My payment failed today." $\rightarrow$ \{Billing Issue\}
    \item For an ambiguous technical problem: "The app crashes when I try to checkout." $\rightarrow$ \{Technical Support, Product Bug\}
    \item For a complex case: "I want to cancel my subscription due to a defective product." $\rightarrow$ \{Cancellation Request, Product Return, Complaint\}
    \end{itemize}
    
This approach ensures reliable routing of customer inquiries, allows for multiple teams to be notified when necessary, and maintains high customer satisfaction by addressing all potential aspects of complex requests.

\end{itemize}


We built confidence intervals with OLS in \autoref{sec:ch29:OLS}. Whilst very useful, these face many limitations spelled out as assumptions in the blue drill-down box. In the sections on random forests (\autoref{sec:ch29:rf}) and neural networks (\autoref{sec:ch29:nn}) that follow, we quantified the uncertainty for individual predictions and showed how to use accuracy-rejection curves for setting uncertainty thresholds to achieve a desired expected accuracy (or any another metric). The uncertainty estimates could also be used to build confidence intervals or sets. But random forests or neural networks may not always be the optimal choice of model, and the derivation of the confidence intervals/set can be rather cumbersome.

Conformal prediction (CF) offers an attractive solution to build confidence intervals and sets \citep{angelopoulos2021gentle}: It works with all data distributions, without any assumptions. CF is model-agnostic: we can use linear regression, random forests, neural networks, or any other prediction model. The trained model suffices: no retraining is required.

There is a tradeoff between the size of the prediction set/interval (ideally small/narrow) and the probability of the true target being contained within it (ideally high): larger sets/intervals will be more likely to contain the true target, and vice versa \citep{straitouri2023improving}. Figures \ref{fig:ch29:OLS_basecase_a} and \ref{fig:ch29:OLS_basecase_b} visualize this tradeoff for the OLS example.

\subsection{Generating predictions sets and intervals}
Conformal prediction transforms any predictive model into one that outputs statistically valid confidence sets (classification) or intervals (regression). It does so by using a held-out calibration set to empirically estimate how uncertain the model is about its predictions, and then forms sets that contain the true answer with a desired probability. Below, we explain the methodology mathematically and then illustrate it with an example. 

We begin by holding out a small part of our training data as a calibration set with $n$ observations and then train the prediction model of our choice on the remaining training set. Given an observation $x_h$ (from the same distribution!\footnote{While CF does not assume a particular distribution or model type, it does require that the calibration and test data are drawn from the same distribution — i.e., they are exchangeable. This assumption is typically considered reasonable when the data is i.i.d. and the test set is drawn from the same source as the training and calibration sets, such as when using a random split of a larger dataset.}), our goal is to create a prediction set $C(x_h)$ (a range or an interval for regression problems) that contains the true value $y_h$ with (average) probability $1-\alpha$:

\begin{equation}\label{eq:ch29:prediction_set}
1-\alpha \leq p(y_h \in C(x_h)) \leq 1-\alpha+\frac{1}{n+1}
\end{equation}

The $\frac{1}{n+1}$ term in the upper bound accounts for the small, unavoidable imprecision inherent when working with a finite calibration set, ensuring the prediction sets still offer a statistically reliable guarantee even with limited data. This finite-sample validity constitutes a major advantage of CF: unlike many other uncertainty quantification methods that rely on asymptotic assumptions (i.e., they only work well with infinite data), CF provides statistically rigorous guarantees even with a finite amount of calibration data.

For a \textbf{classification} problem, we define the conformal score as: 

\begin{equation}\label{eq:ch29:si}
s_i=1-\hat{f}(x_i)_{(y_i)}
\end{equation}

Here, $\hat{f}(x_i)_{(y_i)}$ denotes the predicted probability for the true class label $y_i$ for observation $x_i$ (softmax output for the true target). $s_i$ will be high for bad predictions, and low for good predictions\footnote{For perfect predictions, $\hat{f}(x_i)_{(y_i)}=1$  as the model assigns $100\%$ probability to the true target $y_i$, and none to the other target(s).}.

Next, we determine a threshold $\widehat{q}$ based on these conformal scores. To calculate this value, we first sort the scores $s_i$ for all $n$  observations in the calibration set and get $s_{(1)} \leq s_{(2)} \leq \dots \leq s_{(n)}$. The threshold $\widehat{q}$ is then chosen as the score at rank $k$, where $k$ is calculated as:

$$k=\lceil(n+1)(1-\alpha)\rceil$$

Thus, $\widehat{q} = s_{(k)}$\footnote{The ceiling function $\lceil \cdot \rceil$ ensures that our rank selection is conservative, upholding the coverage guarantee.}. Knowing $\hat{q}$ enables us to construct the prediction set $C(x_h) = \{y:\hat{f}(x_h)_y\ge 1-\hat{q}\}$: we retain all classes $y$ for which the softmax output is greater than or equal to $1-\hat{q}$. The resulting set will satisfy \autoref{eq:ch29:prediction_set}.

The above approach hardly changes for a \textbf{regression} problem. In regression, $\hat{f}(x_i)$ is the model’s point prediction. The conformal score is defined based on how far the true label $y_i$ deviates from this prediction. If we care about absolute errors, we can use:

\begin{equation}
s_i=|y_i-\hat{f}(x_i)|
\end{equation}

Alternatively, we can use:
\begin{equation}
s_i=(y_i-\hat{f}(x_i))^2
\end{equation}

if squared errors are more relevant. The prediction interval around $\hat{f}(x_h)$ has a width determined by $\widehat{q}$.

Because CF only requires softmax scores (classification) or predictions (regression) on a calibration set and test set, it is computationally efficient and does not require any retraining. It scales well with model complexity.

\subsection{Example}

\begin{figure}[htbp]
    \centering
    \includegraphics[width=\textwidth]{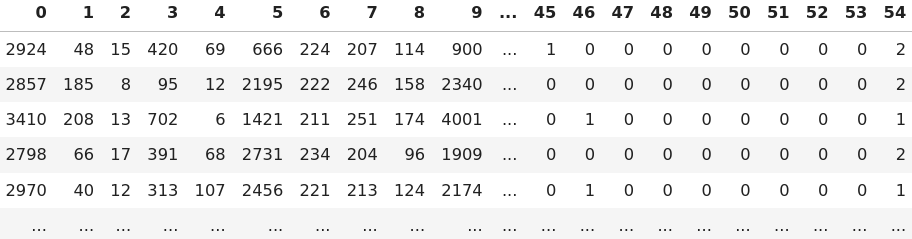}
    \caption{Selected rows and columns from the ‘covertype’ dataset. The last column contains the targets, encoded as integers ranging from 0 to 6. The first columns contain numerical values (elevation, aspect, slope, …), while the latter ones are one-hot encoding columns (soil type and wilderness area).}
    \label{fig:ch29:coverdataset}
\end{figure}

The ‘covertype’ dataset \citep{misc_covertype_31} contains data about four wilderness areas. The task consists of predicting the cover type from cartographic variables. The dataset includes seven class labels or cover types (Spruce/Fir, Lodgepole Pine, Ponderosa Pine, Cottonwood/Willow, Aspen, Douglas-fir, and Krummholz), and 54 independent variables, many of which are one-hot encodings of categorical variables (notably 40 soil types). \autoref{fig:ch29:coverdataset} shows selected rows and columns. We reduced the dataset to 10,000 random observations, split it into a training set (6,400 observations), calibration set (1,600 observations), and test set (2,000 observations), and trained a Random Forest Classifier with 100 estimators and a maximum depth of two.

We now use our trained Random Forest Classifier to produce softmax vectors for all observations in the calibration set, the first five of which are shown in \autoref{tab:ch29:confscores}. The values for the true targets are highlighted in bold and are used to calculate $s_i$ in the last column. Note that these values need not be the highest.

\begin{table}[h]
\centering
\begin{tabular}{|c|ccccccc||c|}
    \hline
    \textbf{Target} & \textbf{0} & \textbf{1} & \textbf{2} & \textbf{3} & \textbf{4} & \textbf{5} & \textbf{6} & $\bm{s_i}$ \\
    \hline
    2 & 0.307 & 0.498 & \textbf{0.109} & 0.004 & 0.024 & 0.031 & 0.028 & 0.891 \\
    1 & 0.286 & \textbf{0.626} & 0.032 & 0.002 & 0.015 & 0.017 & 0.022 & 0.374 \\
    0 & \textbf{0.480} & 0.432 & 0.026 & 0.002 & 0.011 & 0.013 & 0.036 & 0.520 \\
    1 & 0.320 & \textbf{0.542} & 0.052 & 0.003 & 0.021 & 0.029 & 0.029 & 0.458 \\
    0 & \textbf{0.450} & 0.441 & 0.036 & 0.002 & 0.013 & 0.018 & 0.040 & 0.550 \\
    \hline
\end{tabular}
\caption{Softmax predictions for five calibration samples. The first column shows the true class labels (targets). Columns 0–6 represent the softmax vector (predicted class probabilities). The bold value in each row is the predicted probability for the true target. The last column ($s_i$) shows the conformal score for each sample.}
\label{tab:ch29:confscores}
\end{table}

Let’s say $\alpha=20\%$; in other words, we want to construct sets that contain the true targets with 80\% probability. When we sort the vector of $[s_1, ..., s_{1600}]$ and scan for the $\lceil(1,600+1)(1-.2)\rceil =  1,281$-th quantile $\hat{q}$, we find $\hat{q}=0.645$.

Let’s now compute the softmax vectors for all test set observations using our trained Random Forest Classifier. The first five vectors are shown in \autoref{tab:ch29:softtest}.

\begin{table}
\begin{center}
\begin{tabular}{|c|c|c|c|c|c|c|}
    \hline
    \textbf{0} & \textbf{1} & \textbf{2} & \textbf{3} & \textbf{4} & \textbf{5} & \textbf{6} \\
    \hline
    0.337 & \textbf{0.529} & 0.053 & 0.003 & 0.019 & 0.028 & 0.031 \\
    0.175 & \textbf{0.362} & 0.269 & 0.026 & 0.014 & 0.137 & 0.016 \\
    0.204 & \textbf{0.362} & 0.269 & 0.026 & 0.013 & 0.124 & 0.018 \\
    \textbf{0.435} & \textbf{0.444} & 0.044 & 0.003 & 0.014 & 0.021 & 0.039 \\
    \textbf{0.434} & \textbf{0.390} & 0.030 & 0.002 & 0.012 & 0.015 & 0.117 \\
    \hline
\end{tabular}
\caption{Softmax predictions on the test set. Bold values (every class $y$ for which $\hat{f}(x_h)_y \geq 1 - \hat{q}$) are retained in the final prediction set.}
\label{tab:ch29:softtest}
\end{center}
\end{table}

To build the final prediction sets, we need to check which softmax probabilities in \autoref{tab:ch29:softtest} are larger than or equal to $1-\hat{q} = 0.355$. These are represented in bold font. The final prediction sets for the first five observations in the test set thus become: \{1\}, \{1\}, \{1\}, \{0, 1\}, and \{0, 1\}. When we compare the 2,000 prediction sets thus generated for the complete test set with the corresponding true targets, we find that the true targets are part of the prediction set in 80.35\% of the cases, which is very close to the required 80\%.

\section{Working with Uncertainty}
In this chapter, you gained an understanding of machine learning-related uncertainty. This understanding confers several benefits when thinking about or working with prediction systems. We discuss these benefits in this section.

\subsection{Improving prediction systems}
An understanding of machine learning-related uncertainty can improve prediction systems even before they are developed or bought. The machine learning pipeline starts with data. The concepts of epistemic and aleatoric uncertainty provide a framework for assessing whether the availability and quality of that data are sufficient. This understanding also helps inform decisions about what kinds of data should be recorded or acquired to support future prediction tasks.

A prediction system or model’s ability to deliver uncertainty estimates permits the interpretation of its predictions (\autoref{subsec:interpretpred}) and facilitates automation (\autoref{subsec:autom}). This ability should therefore be an important decision criterion when selecting a candidate system/model for a given prediction task. 

Once the predictions system is in production, high epistemic or aleatoric uncertainty, overall or in certain parts of the domain, can provide useful clues for improvement.

High \textit{aleatoric uncertainty}, one the one hand, can point to data quality problems. Sensor errors can be caused by hardware or software problems, environmental conditions, interference, improper installation, etc. When working with data annotated by human labelers — including not only professional annotators but also implicit signals from customers, such as clicking like-buttons, sharing or forwarding messages, or making purchases after receiving a voucher — human error or bias is always possible and demands careful investigation.
When essential explanatory variables are missing, a model may exhibit high aleatoric uncertainty because the labels appear random from the model’s perspective. This often occurs in specific subgroups within the data. For example, imagine an online retailer trying to predict whether a customer will return an item. The model might perform well for customers in most regions but poorly in a particular country. Upon investigation, it turns out that this country has a unique return policy — such as free return shipping or extended return windows — but this information was not included in the model. From the model’s point of view, customer returns in this subgroup appear unpredictable or noisy because it's missing a key variable that influences return behavior. Once this return policy data is incorporated as a feature, the model gains explanatory power, and aleatoric uncertainty is reduced.

High \textit{epistemic uncertainty}, on the other hand, can point to a lack of data in certain parts of the domain. We learned that epistemic uncertainty is reducible: more data will help. However, more data comes at a cost: either it needs to be acquired (e.g., weather data) from external sources or collected by recording and storing it internally. Data collection may require the installation of hardware or involve experiments with customers. This trade-off between the utility and the cost of additional data is the subject of active learning, a field where a model's epistemic uncertainty is used to direct the acquisition of new data, akin to a student asking questions to a teacher.

In a running system, rising uncertainty can also signal data drift—changes in the input distribution—or concept drift, where the relationship between inputs and outputs evolves over time. Monitoring uncertainty can thus support early detection of such shifts, prompting model retraining or data updates before performance further degrades.

\subsection{Interpreting predictions}\label{subsec:interpretpred}
The running production system will generate predictions that uncertainty estimates can help interpret. Earlier in this chapter, we described three ways how this can happen with thresholds, confidence intervals, and conformal prediction.

\begin{figure}[htbp]
    \centering
    \includegraphics[width=.4\textwidth]{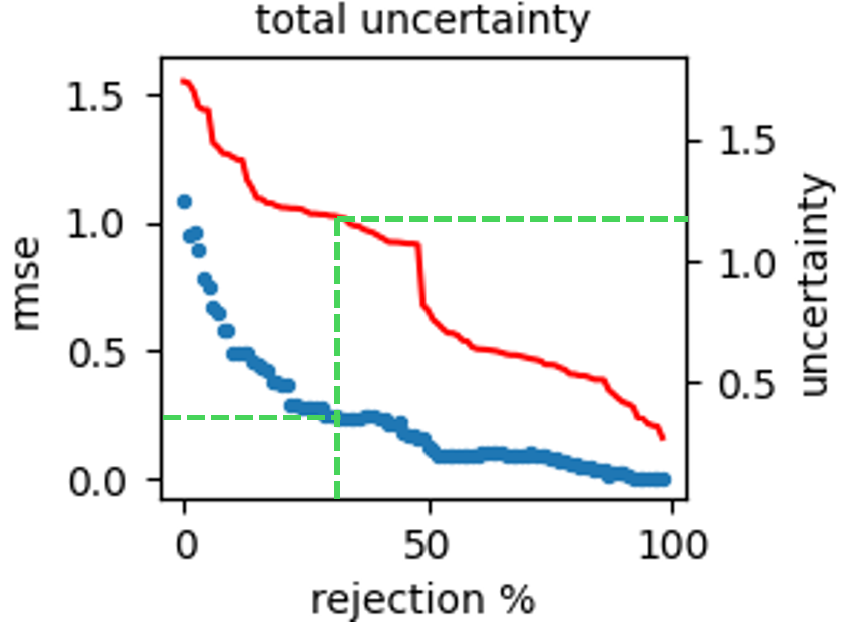}
    \caption{The accuracy-rejection curves for the total uncertainty from \autoref{fig:ch29:acc_rej_BNN} used for setting a threshold: A required RMSE of 0.25 demands a total uncertainty threshold of 1.2.}
    \label{fig:ch29:threshold}
\end{figure}

In our discussion of accuracy-rejection curves (\autoref{subsec:accrej}), we demonstrated that uncertainties correlate strongly with the accuracy of predictions, enabling organizations to set arbitrary accuracy \textit{thresholds} for their prediction systems. This can be achieved by computing the corresponding uncertainty threshold and retaining only those predictions with uncertainties that fall below it, as illustrated in \autoref{fig:ch29:threshold}. Predictions with higher uncertainty are rejected and can be passed on to humans or another system.

\textit{Confidence intervals} offer a statistical approach to quantify uncertainty, helping mitigate the risks associated with unreliable predictions. They can also be visualized to communicate the range of plausible outcomes, aiding interpretation and decision-making. 

Conversely, \textit{conformal prediction} provides a flexible and model-agnostic framework for generating prediction sets that meet predefined confidence levels, offering a practical way to express uncertainty even in classification tasks.

\subsection{Facilitating automation}\label{subsec:autom}
\paragraph{Human-machine collaboration}
Awareness of prediction uncertainty enables the design of dual-track systems, where high-confidence cases proceed through automated processing, while low-confidence cases (those exceeding an uncertainty threshold) are routed to human experts. These uncertain cases are often more complex, rare, or ambiguous, making them better suited for human judgment. Such a division of labor can enhance job satisfaction and ensure more meaningful use of human expertise.

\paragraph{Working with smaller datasets: earlier adoption of prediction systems}
Our discussion of epistemic uncertainty showed that insufficient data leads to inaccurate predictions. Consequently, organizations are often unable to deploy prediction systems until adequate training data becomes available.  This delay can be costly in the current information age, characterized by rapid innovation, fast implementation, and continuous learning. By setting appropriate uncertainty thresholds, organizations can begin using prediction systems earlier. Initially, only a small proportion of predictions may meet the threshold for acceptance, but this proportion will grow as more data is collected. During this phase-in period, organizations gain valuable insights that can enhance both system performance and data collection strategies.

\bibliography{bibliography}

\end{document}